\documentclass[10pt,twocolumn,letterpaper]{article}

\usepackage{cvpr}              

\usepackage{graphicx}
\usepackage{amsmath}
\usepackage{amssymb}
\usepackage{booktabs}
\usepackage{cite}
\usepackage{times}
\usepackage{epsfig}
\usepackage{graphicx}
\usepackage{amsmath}
\usepackage{amssymb}
\usepackage{xcolor}
\usepackage{multirow}
\usepackage{makecell}
\usepackage{algorithm}
\usepackage{algorithmic}
\usepackage{bbm}
\usepackage{url}
\usepackage{adjustbox}
\usepackage{threeparttable}
\usepackage{caption}
\usepackage[pagebackref,breaklinks,colorlinks]{hyperref}
\hypersetup{
	colorlinks=true,
	linkcolor=red,
	filecolor=red,
	citecolor=green,      
	urlcolor=cyan,
}

\usepackage[capitalize]{cleveref}
\crefname{section}{Sec.}{Secs.}
\Crefname{section}{Section}{Sections}
\Crefname{table}{Table}{Tables}
\crefname{table}{Tab.}{Tabs.}

\def\cvprPaperID{11} 
\def\confName{CVPR}
\def\confYear{2023}

\begin{document}

\title{Correlation Pyramid Network for 3D Single Object Tracking}
\author{Mengmeng Wang$^{1}$, Teli Ma$^{1}$, Xingxing Zuo$^{2}$, Jiajun Lv$^{1}$, \vspace{0.3cm} Yong~Liu$^{1*}$ \\
   $^1$Zhejiang University
 \quad 
  $^2$Technical University of Munich \\
\texttt{\{mengmengwang, lvjiajun314\}@zju.edu.cn},\quad
\texttt{telima9868@gmail.com}\\
\texttt{xingxing.zuo@tum.de},\quad
\texttt{yongliu@iipc.zju.edu.cn}
\thanks{Yong~Liu is the corresponding author.}
}

\maketitle

\begin{abstract}
 3D LiDAR-based single object tracking (SOT) has gained increasing attention as it plays a crucial role in 3D applications such as autonomous driving. The central problem is how to learn a target-aware representation from the sparse and incomplete point clouds. In this paper, we propose a novel Correlation Pyramid Network (CorpNet) with a unified encoder and a motion-factorized decoder. Specifically, the encoder introduces multi-level self attentions and cross attentions in its main branch to enrich the template and search region features and realize their fusion and interaction, respectively. Additionally, considering the sparsity characteristics of the point clouds, we design a lateral correlation pyramid structure for the encoder to keep as many points as possible by integrating hierarchical correlated features. The output features of the search region from the encoder can be directly fed into the decoder for predicting target locations without any extra matcher. Moreover, in the decoder of CorpNet, we design a motion-factorized head to explicitly learn the different movement patterns of the up axis and the x-y plane together. Extensive experiments on two commonly-used datasets show our CorpNet achieves state-of-the-art results while running in real-time.
\end{abstract}
\vspace{-0.9cm}
\section{Introduction}
\label{sec:intro}
This paper focuses on 3D LiDAR single object tracking (SOT), which is emerged in recent years but is an essential task for 3D applications like autonomous driving, robotics, and surveillance system with the development of 3D sensors like LiDAR. The task aims at tracking a specific target in a video by giving the corresponding 3D target bounding box in the first frame. It is challenging since the target will undergo several changes like occlusions and fast motions, and be sparse and incomplete.
\begin{figure}[htbp]
		\begin{center}
			\includegraphics[width=1\linewidth]{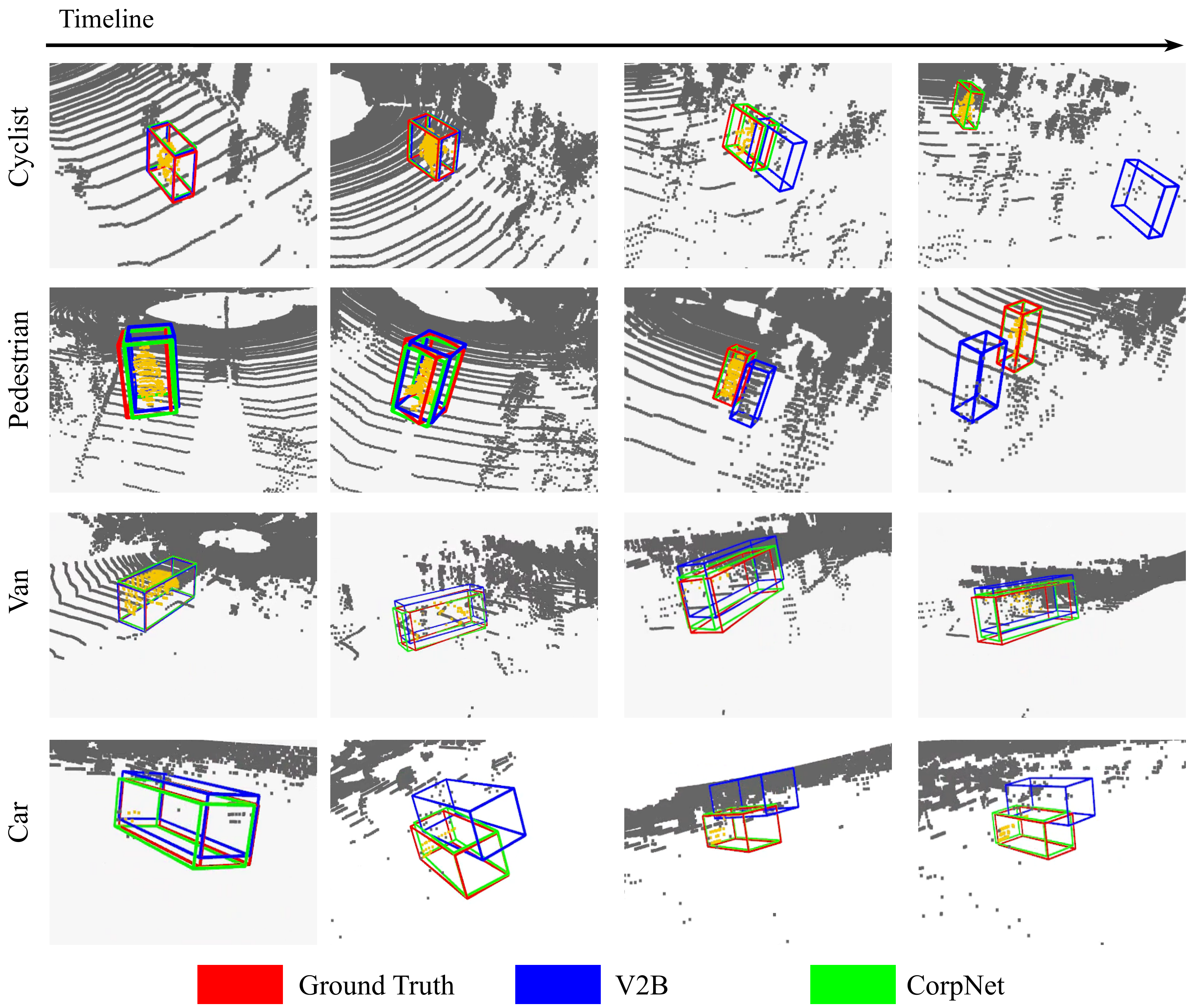}
		\end{center}
  \vspace{-0.5cm}
		\caption{Visualization results of the four different categories. The points of the targets are highlighted in \textcolor{yellow}{yellow}. The \textcolor{red}{red} boxes are ground truth bounding boxes. The \textcolor{green}{green} boxes are the objects tracked by our CorpNet, while the \textcolor{blue}{blue} boxes are the results of V2B. }
		\label{vis}
		\vspace{-0.5cm}
\end{figure}

Prior arts mainly addressed the above challenges with a conventional \textit{extractor-matcher-decoder} paradigm. The extractor is used to encode the features of the template and the search areas. The matcher is employed to build the template-search relationship and enhance the potential target features, i.e., embedding the template features into the features of the search region, which is also called correlation operation. The decoder is leveraged to generate bunches of 3D target proposals based on the features from the matcher.
Since the modern backbones \cite{pointnet,qi2017pointnet} have become the mainstream and even default choices for the extractor, most trackers in this paradigm dedicate to design more robust and elaborate matchers \cite{qi2020p2b,zheng2021box,hui20213d,zhou2022pttr,cui20213d}, and more powerful decoders \cite{qi2020p2b,hui20213d}. Despite their great success, we find they still suffer from two limitations. (1) The relationship between the template and the search features is always modeled only in the matcher, which is not sufficient for completely cross-source interaction and target enhancement. (2) The point downsampling in the commonly used backbones will inevitably exacerbate the sparsity of the point clouds.

The first limitation is mainly caused by the conventional paradigm structure, which separates the extractor and matcher, and makes the matcher be responsible for the template feature enhancement. Inspired by the 2D SOT methods \cite{bertinetto2016fully,li2018high,dong2019quadruplet,huang2019got,he2018twofold,zhang2019deeper,xu2020siamfc,lu2020deep}, previous methods in this paradigm \cite{giancola2019leveraging,qi2020p2b,zheng2021box,shan2021ptt,zhou2022pttr,hui20213d,fang20203d,wang2021mlvsnet,stnet} always employ a Siamese-like backbone in the extractor that independently embeds the template and search frames without any intermediate interaction. They then design various matchers to fuse the template features into the search features. However, a standalone matcher is redundant, and the extracted high-level features used in the matcher are not sufficient. $M^2$-Track \cite{zheng2022beyond} has realized part of this problem and proposed a motion-centric paradigm to avoid the Siamese-like backbone and predicts the motion directly. Nevertheless, they still need a motion transformation module to integrate the template information into the search representations, and a two-stage refinement is used to ensure the performance. Inspired by the recent trend in 2D SOT~\cite{xie2022correlation,ostrack2022,simtrack2022,mixformer2022}, 
our key insight is that the extractor should be responsible for feature representation and matching simultaneously, so no extra matcher is needed. 

For the second limitation, we have noticed that the backbone of previous extractors in 3D SOT remains unexplored. The default configuration is the Siamese-like PointNet \cite{pointnet}/PointNet++ \cite{qi2017pointnet}, which are not originally designed for 3D SOT. For example, the most commonly-used PointNet++ usually downsamples the input points (typically 1024 or 512 points) by 8$\times$, remaining fewer points (128 or 64 points) for matcher and decoder, and making the intermediate features from the extractor even much sparser than the already sparse input. Actually, the feature embedding backbone plays a core role in object tracking but is overlooked in the previous 3D trackers. It needs to provide a discriminative target representation of the input sparse point clouds, with the around background that inevitably includes distractors and noises. To ease this problem, PTTR \cite{zhou2022pttr} proposes a relation-aware sampling strategy for preserving more template-relevant points. However, the total preserved points are still sparse, so they need a two-stage refinement to keep the performance. We emphasize this problem from another perspective, which is to adjust the architecture of the backbone and keep the points from all the stages to formulate multi-scale representations, strengthening the representational capacity of the proposed framework.

Considering the above two limitations, we propose a novel \textbf{Cor}relation \textbf{P}yramid \textbf{Net}work (CorpNet). To be specific, the encoder of CorpNet introduces Self Attention (SA) and Cross Attention (CA) modules in multiple stages of the backbone, strengthening the frame representation at multiple levels by SA and facilitating sufficient interaction to replace the original matcher by CA. 
Afterward, to cope with the sparsity brought by the downsampling operation, we formulate a correlation pyramid architecture in the encoder to reserve as many points as possible. More specifically, a lateral correlation pyramid structure is devised to effectively combine the point features from all the stages, which have different amounts of points and feature dimensions. Then the pyramidally fused features are voxelized to a volumetric representation and fed to the decoder. The main branch of the encoder deeply embeds the cross-source feature in multiple layers, and the lateral correlation pyramid extensively and directly combines correlated features from low level to high level, resulting in sufficient target-aware feature extraction. Moreover, our CorpNet builds a new decoder based on \cite{hui20213d} to process the powerful representation from the encoder, considering that the motion of the \textit{x-y} plane and the \textit{z} axis are not exactly identical. 
Finally, as shown in Fig. \ref{vis}, with the proposed CorpNet, we achieve state-of-the-art performance on two widely adopted datasets (i.e., KITTI \cite{kitti} and NuScenes \cite{nuscenes}).

The main contributions of our paper can be summarized as follows:
\begin{itemize}
	\item We propose a novel \textbf{Cor}relation \textbf{P}yramid \textbf{Net}work dubbed CorpNet, which integrates multi-level self attentions to enrich the representations and multi-scale cross attentions to empower sufficient interaction/matching between them. 
	\item We emphasize the sparsity problem of the downsampling operation by a new correlation pyramid structure that fuses the hierarchical correlated features to reserve features of all the points inside different stages. 
	\item We design a new motion-factorized decoder to explicitly decouples the prediction of the \textit{x-y} plane and \textit{z} axis for their different motion patterns.
\end{itemize}

\section{Related Works}\label{sec:related}
3D object tracking \cite{ong2020bayesian,crivellaro2017robust,hu2022monocular,giancola2019leveraging,zhang2022voxeltrack} works with 3D input data like point clouds, stereo images and even monocular images. Here we discuss the task of 3D single object tracking task. This is a new task that has emerged in recent years, which is first defined in SC3D \cite{giancola2019leveraging}. Inspired by the structure of 2D SOT, previous methods \cite{giancola2019leveraging,qi2020p2b,fang20203d,zarzar2019efficient,cui20213d,shan2021ptt,zhou2022pttr,zheng2021box,hui20213d}, of 3D SOT all inherit a \textit{extractor-matcher-decoder} paradigm. As a pioneer, SC3D lays the foundation of this paradigm with simple components, which matches cosine similarity (matcher) of features between candidates and target (Siamese-like extractor) and regularizes the training using shape completion (decoder). 
\begin{figure*}[tbp]
		\begin{center}
			\includegraphics[width=1\linewidth]{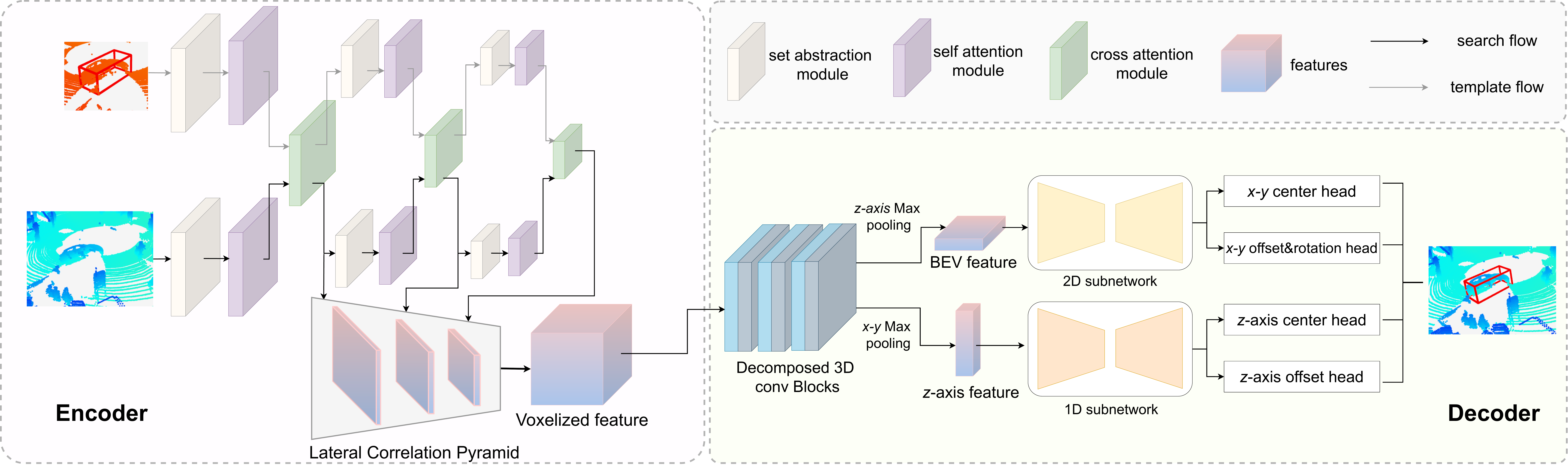}
		\end{center}
    \vspace{-0.6cm}
		\caption{The overall architecture of the proposed CorpNet. Given template and search point clouds and the BBox of the template, a unified encoder is used to extract the representations and match the template and search area simultaneously. A lateral correlation pyramid structure is proposed to handle the sparsity and incomplete challenges by leveraging the hierarchical correlated features. Then a decomposed decoder is designed to obtain the motion of the \textit{x-y} plane and \textit{z}-axis separately. The whole pipeline is end-to-end trained with only a single stage.}
		\label{overall}
		\vspace{-0.4cm}
\end{figure*}

The following trackers mainly focus on improving SC3D from two folds. The first dedicates to design more robust matchers \cite{qi2020p2b,zheng2021box,hui20213d,zhou2022pttr,cui20213d,wang2021mlvsnet,zheng2022beyond}. For instance, MLVSNet \cite{wang2021mlvsnet} uses the CBAM module \cite{woo2018cbam} to enhance the vote cluster features with channel attentions and spatial attentions. V2B \cite{hui20213d} employs the global and local template feature embedding to strengthen the correlation between the template and search area. $M^2$-Track introduces a motion-centric paradigm and uses input merge and motion transformation to combine the template and search features instead of conventional correlation operation in the matcher. However, a two-stage refinement is needed to ensure the performance for the lack of appearance matching. The prior effort in the matcher still struggles with standalone matchers, which could not explicitly benefit the extractor and make the structures redundant.
In the second fold, the trackers \cite{qi2020p2b,fang20203d,hui20213d,wang2021mlvsnet,shan2021ptt} have tried to improve the decoder part. P2B \cite{qi2020p2b} employs Hough Voting to predict the target location and many methods follow \cite{zheng2021box} or improve \cite{shan2021ptt,wang2021mlvsnet} this manner. LTTR \cite{fang20203d} and V2B \cite{hui20213d} use center-based regressions to predict several object properties. Even though the matcher and decoder are explored a lot, we find the extractor is always neglected since the modern backbones \cite{pointnet,qi2017pointnet} become the mainstream and default choice. We figure out that the extractor is crucial for a powerful representation to play as the fundamental of the matcher and decoder. Therefore, we shed light on this point and propose a new single-stage correlation pyramid network to explore a unified backbone specified for the 3D SOT task and merge the extractor and matcher together.

\section{Method}\label{sec:method}
\subsection{Problem Definition}
The 3D LiDAR single object tracking (SOT) task is defined as: Given a dynamic 3D sequence of $T$ point clouds $\{\mathcal{P}_{i} \}_{i=1}^{T}$ and an initial bounding box (BBox) $\mathcal{B}_{1}=(x_{1},y_{1},z_{1},w_{1},l_{1},h_{1},\theta_{1})$ of a target, our goal is to localize the target BBoxs $\{\mathcal{B}_{i} \}_{i=2}^{T}$ in sequential frames online, where the subscript stands for the sequence id, $(x,y,z)$ indicates the center coordinate of the BBox, $(w,l,h)$ is the BBox size and $\theta$ is the heading angle (the rotation around the \textit{up}-axis). Generally, the BBox size is assumed to keep unchanged across all frames in the 3D scenes even for non-rigid objects
(the BBox size for a non-rigid object is defined by its maximum extent in the scene), so we do not need to re-predict the size and simplify $\mathcal{B}_{i}$ from $\mathbb{R}^{7}$ to $\mathbb{R}^{4}$.
\begin{figure*}[tbp]
		\begin{flushright}
     \centering
			\includegraphics[width=1\linewidth]{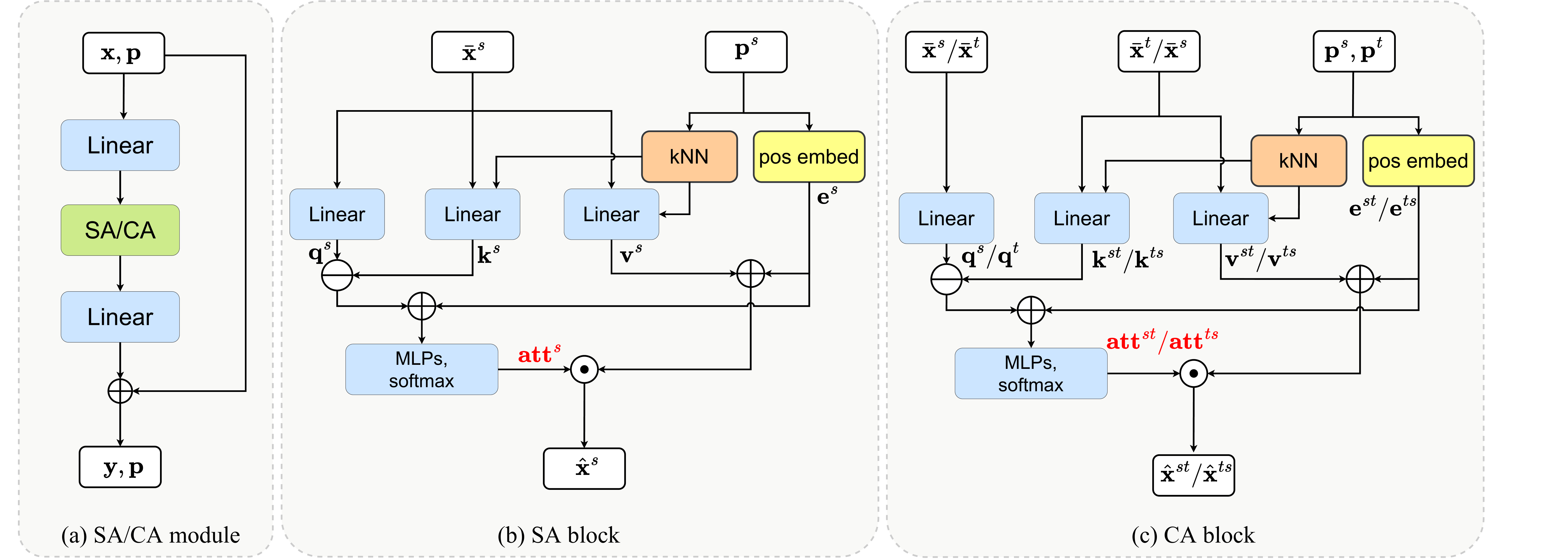}
		\end{flushright}\vspace{-0.5cm}
		    \caption{Illustration of the SA/CA module and the SA/CA blocks of the encoder. ``att" is short for attention.}
		\label{saca}
		\vspace{-0.5cm}
\end{figure*}

More specifically, we follow the conventional tracking pipelines to generate the input consisting of a template area $\mathcal{P}^t=\{\textbf{P}_{i}^{t}\}_{i=1}^{N_t}$ and a search area $\mathcal{P}^s=\{\textbf{P}_{i}^{s}\}_{i=1}^{N_s}$, where $\textbf{P}_{i}\in \mathbb{R}^{3+d}$ is a 3D $(x,y,z)$ point with $d$-dimensional features like intensity and elongation. ${N_t}$ and ${N_s}$ are the point number of the template and search areas, respectively. The template $\mathcal{P}^t$ is cropped and centered in the template frames with its corresponding BBox and the search region $\mathcal{P}^s$ is cropped and centered according to $\mathcal{B}_{k-1}$ with an enlarged area in the search frame. 

\subsection{CorpNet}
The overall architecture of the proposed CorpNet is shown in Fig. \ref{overall}, which consists of a unified encoder and a factorized decoder.
\subsubsection{Encoder} \label{encoder}
The function of the encoder is not only to extract features for the template and search regions but also to simultaneously be responsible for the feature matching or correlation. The popular Siamese-like backbones are not suitable anymore in this situation. We design a unified structure to satisfy the requirements. 

As shown in Fig. \ref{overall}, the encoder of our CorpNet consists of three stages, where each of them contains a set abstraction module \cite{qi2017pointnet} to gradually reduce the point number, a self attention (SA) module to enrich the feature representations and a cross attention (CA) module to implement feature correlation and interaction. Specifically, the set abstraction modules select a set of points from input points, then group them with the local neighborhood of a ball query and embed the local region patterns into feature vectors with pointwise MLPs (we only use two MLP layers here). Next, each stage's core SA and CA are detailedly introduced as follows.

Most attention mechanisms \cite{guo2021pct,dosovitskiy2020image} are global attentions with scalar dot products. It is known that the computation of global attention over all input tokens leads to quadratic complexity. 
However, the inference speed should be considered in real-time applications for the 3D SOT task. Therefore, a local vectorized self-attention mechanism \cite{zhao2021point} is leveraged in CorpNet to diminish the computational overhead.
In other words, for one specific point, the attention is calculated from several adjacent points around it rather than all the points. 

Let $\mathbf{x}_i,\mathbf{p}_i$ denote the feature, location of the $i$-th point, respectively. To illustrate the commonalities of SA and CA modules, we omit the superscript $s/t$ of the search/template region here. A SA/CA module (Fig. \ref{SA_CA}(a)) consists of a linear layer $f_1$, a SA/CA block, another linear layer $f_2$ and a residual connection. Formally, 
\begin{equation}\label{sacablock}
\begin{aligned}
    \bar{\textbf{x}}_i&=f_1(\mathbf{x}_i),\\
    \hat{\textbf{x}}_i&=\texttt{SA/CA}(\bar{\textbf{x}}_i),\\
    \textbf{y}_i&=f_2(\hat{\textbf{x}}_i)+\mathbf{x}_i \text{.}
\end{aligned}
\end{equation}

In the SA/CA block (Fig. \ref{SA_CA}(b/c)), we obtain the query, key and value ($\textbf{q}_i,\textbf{k}_i,\textbf{v}_i$) features as:
\begin{equation}
\begin{aligned}
    \textbf{q}_i&=f_q(\bar{\mathbf{x}}_i),\\
    \textbf{k}_i&=f_k( \texttt{kNN}(\mathbf{p}_i),\bar{\mathbf{x}}_i),\\
    \textbf{v}_i&=f_v(\texttt{kNN}(\mathbf{p}_i),\bar{\mathbf{x}}_i)
\end{aligned}
\end{equation}
where $f_q$, $f_k$, $f_v$ are linear projections and \texttt{kNN} denotes the $k$ nearest neighbors. The position encoding is defined as:
\begin{equation}
\begin{aligned}
    \textbf{e}_{ij}&=f_e(\textbf{p}_i-\textbf{p}_j)
\end{aligned}
\end{equation}
where $\textbf{p}_i$ and $\textbf{p}_j$ are the 3D position for points $i$ and $j$. $f_{e}$ is a linear projection. Then the core local attention mechanism could be formulated as:
\begin{equation}
\begin{aligned}
    \hat{\mathbf{x}}_i=\sum_{j=1}^{k}\sigma (\frac{1}{\sqrt{k}}f_{a}(\mathbf{q}_{i}&-\mathbf{k}_{ij}+\mathbf{e}_{ij}))\odot (\mathbf{v}_{ij}+\mathbf{e}_{ij}))
\end{aligned}
\end{equation} 
where $k$ stands for the $k$ nearest neighbors, $f_{a}$ is a two-layer MLPs, $\odot$ represents the element-wise multiplication and $\sigma$ is a softmax function. As illustrated in Fig. \ref{saca}(b), in SA module, both $\texttt{kNN}$ and ($\textbf{q}_i,\textbf{k}_i,\textbf{v}_i$) are calculated from a same source (the template or search region), and the output of the SA block of Eq. \ref{sacablock} is modified to:
\begin{equation}\label{SA}
\begin{aligned}
    \mathbf{y}_i^s&=f_2(\hat{\textbf{x}}_i^s)+\mathbf{x}_i^s,\\
    \mathbf{y}_i^t&=f_2(\hat{\textbf{x}}_i^t)+\mathbf{x}_i^t
\end{aligned}
\end{equation}
where $s$ stands for search region and $t$ is the template region. While in the CA module (Fig. \ref{saca}(c)), $\texttt{kNN}$ is calculated with different sources. Also, $\textbf{q}_i$ and ($\textbf{k}_i,\textbf{v}_i$) are calculated crossly to interact between the template and the search regions and absorb useful target-related features. More specifically, we have:
\begin{equation}\label{CA}
\begin{aligned}
    \mathbf{y}_i^{s}&=f_2(\hat{\textbf{x}}_i^{st}) + \mathbf{x}_i^s,\\
    \mathbf{y}_i^{t}&=f_2(\hat{\textbf{x}}_i^{ts}) + \mathbf{x}_i^t.
\end{aligned}
\end{equation}
From Eq. \ref{SA} and Eq. \ref{CA}, it can be seen that the original features are aggregated with the corresponding transformed/attended one in SA and sufficiently interacted with each other source in CA. 
By stacking multiple stages, the features of both the template and search regions are gradually concentrated on the beneficial target-relevant features.

\subsubsection{Lateral Correlation Pyramid} \label{sec:pyramid}
\begin{figure}[tbp]
	\begin{center}
		\includegraphics[width=0.9\linewidth]{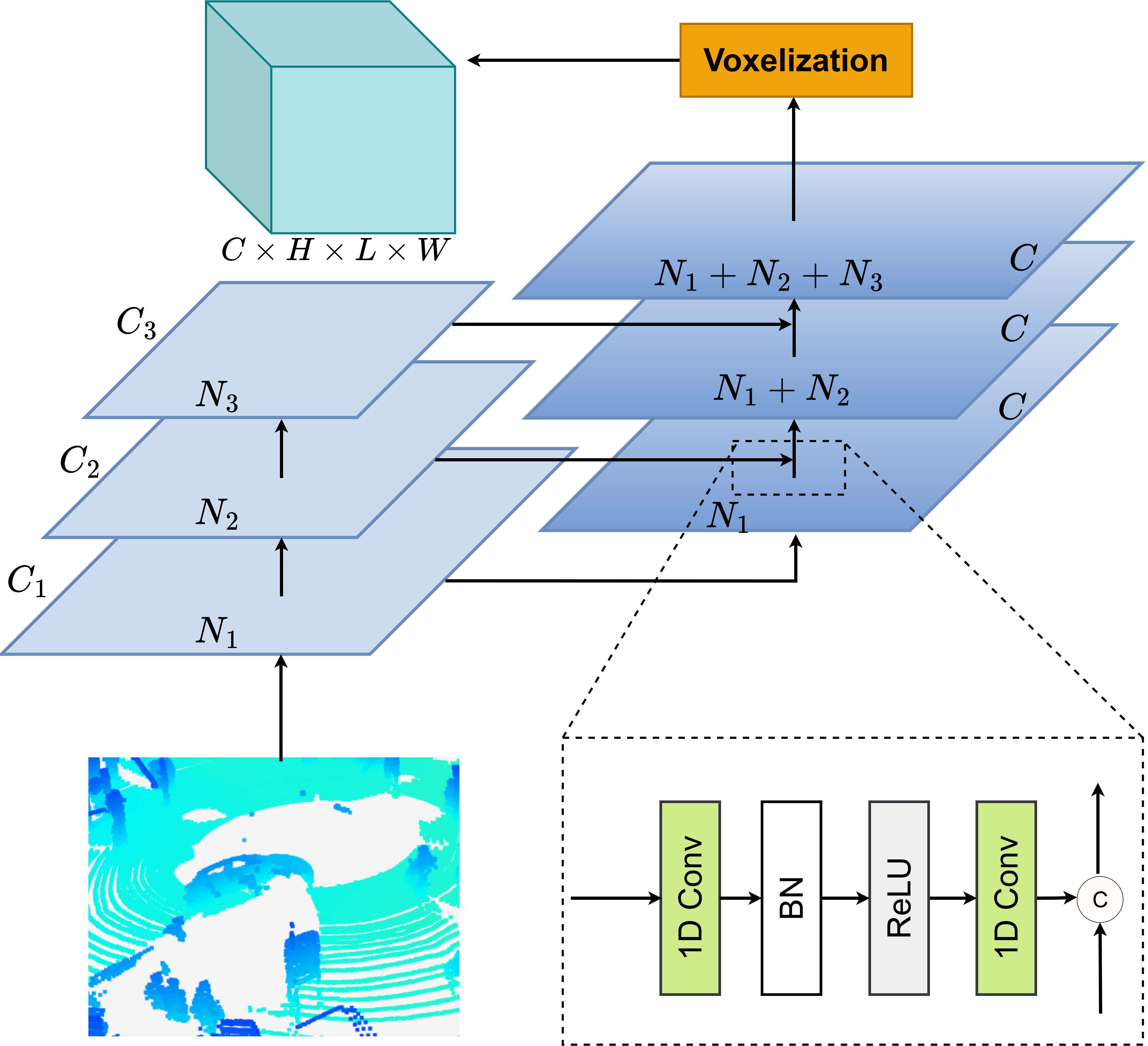}
	\end{center}\vspace{-0.4cm}
	\caption{The structure of the lateral correlation pyramid. The left feature maps are obtained by the three stages of the main branch in the encoder. On the right is the lateral correlation pyramid, which combines every correlated feature, feeding through a 1D convolution block and then merging with the features of the last level by concatenation.}
	\label{pyramid_structure}
			\vspace{-0.5cm}
\end{figure}

In the traditional \textit{extractor-matcher-decoder} paradigm, researchers always overlook the importance of the extractor, which always has several downsampling operations \cite{pointnet,qi2017pointnet} that are used to reduce the model size but exacerbate the sparsity problem. Besides, due to the separation of the extractor and matcher, using the highest-level to do the matching operation in the matcher is a default configuration \cite{hui20213d,qi2020p2b,zheng2021box,zhou2022pttr,fang20203d}, which hinders useful multi-scale combination. Differently, we break the above limitations and try to explore a pyramid structure to equip our CorpNet with rich semantics at all feature levels, as shown in Fig. \ref{pyramid_structure}. 

The feature pyramid is exploited in 2D detection \cite{lin2017feature} but remains unexplored in 3D SOT. In particular, to compensate for the exacerbated sparsity caused by the downsampling of the set abstraction module and keep as many points as possible, we formulate a correlation pyramidal encoder by leveraging the encoder’s pyramidal feature hierarchy, which has semantics from low to high levels. The inherent multi-scale correlated features output from the CA modules of all stages are fused together. Note that the point amount of these features is different due to the downsampling of the set abstraction modules. Also, the feature dimension and the semantic level are different. To this end, for the feature of each stage $\textbf{F}^k \in \mathbb{R}^{N_k \times C_k}$, $k\in \{1,2,3\}$, we first unify the feature dimension by a 1D convolution block which consists of a 1D convolution layer, a BN layer, a ReLU activation, and another 1D convolution layer. Then the obtained features are concatenated with the last features of the pyramid along the point amount dimension. The output features of the lateral pyramid $\textbf{F} \in \mathbb{R}^{(N_1+N_2+N_3) \times C}$ will then be voxelized as a volumetric representation $\mathbf{F}^{m}\in \mathbb{R}^{C\times H\times L\times W}$ by averaging the 3D coordinates and features of the points in the same voxel bin. Given the voxel size $(v_x,v_y,v_z)$ and the range of the search region $[(x_{min},x_{max}),(y_{min},y_{max}),(z_{min},z_{max})]$, the resolution $(W,L,H)$ of the $\mathbf{F}^{m}$ is:
\begin{equation}\label{whl}
\begin{aligned}
W&=\lfloor \frac{x_{max}-x_{min}}{v_x} \rfloor+1,\\
L&=\lfloor \frac{y_{max}-y_{min}}{v_y} \rfloor+1,\\
H&=\lfloor \frac{z_{max}-z_{min}}{v_z} \rfloor+1
\end{aligned}
\end{equation}
where $\lfloor \cdot \rfloor$ indicates the floor operation.
Since only the search region representations are fed into the decoder for prediction in our CorpNet, thus the lateral correlation pyramid serves only for the multi-level embeddings of the search region.

\subsubsection{Decoder}
Since the obtained encoded feature $\mathbf{F}^{m}\in\mathbb{R}^{C\times H\times L\times W}$ is a voxel representation, we can take advantage of regular convolutions that are commonly used in the images. The most similar decoder is from \cite{hui20213d}, which first uses 3D convolutions on their encoded features and pools a BEV feature map to regress the final results, including the \textit{z}-axis location.

In fact, the motion pattern of the \textit{x-y} plane and the movement in \textit{z} axis are not exactly the same. Considering this, we decide to separately model them in our decoder as presented in the right part of Fig. \ref{overall}. First, we disentangle the 3D convolution block into a 2D convolution block (consisting of a 2D convolution layer, a batch normalization layer and a ReLU activation layer) and a 1D convolution block (consisting of a 1D convolution layer, a batch normalization layer, and a ReLU activation layer), operating successively on the BEV and the vertical direction. 
We can further gain another advantage that is an additional nonlinear rectification between these two operations.
This effectively doubles the number of nonlinearities compared to a single 3D convolutions block, thus rendering the model capable of representing more complex functions. After stacking several decomposed 3D convolution blocks, we apply two max pooling operations on the \textit{z} axis and \textit{x-y} plane to get the BEV features $\mathbf{F}^{BEV}$ and vertical features $\mathbf{F}^{z}$, respectively. Mathematically, 
\begin{equation}\label{pool}
\begin{aligned}
    \mathbf{F}^{BEV}&=\texttt{MaxPool}_{z}\phi (\mathbf{F}^{m}),\\
    \mathbf{F}^{z}&=\texttt{MaxPool}_{xy}\phi (\mathbf{F}^{m})
    \end{aligned}
\end{equation}
where $\phi$ is the decomposed 3D convolution blocks. The subscript of the $\texttt{MaxPool}$ means applying max pooling on the corresponding dimension. 

Next, different to \cite{hui20213d,stnet}, we employ two separate prediction subnetworks on top of the $\mathbf{F}^{BEV}$ and $\mathbf{F}^{z}$. Each subnetwork applies a stack of common convolutions blocks (2D convolution for BEV feature maps and 1D convolution for vertical features) on the dense feature map ($\mathbf{F}^{BEV}$ and $\mathbf{F}^{z}$) to strengthen and adjust the features for sufficient local information in the corresponding feature maps. Then, for each subnetwork, a center classification head is attached to predict the discrete object center. To compensate for the discretization error, we regress the offset of the continuous ground truth center with an offset regression head. The rotation regression is predicted together with the BEV offset regression head. The final training objective is expressed as:
\begin{equation}\label{head}
\begin{aligned}
    \mathcal{L}=\lambda_{cls}(\mathcal{L}_{cls}^{BEV}+\mathcal{L}_{cls}^{z})+\lambda_{reg}(\mathcal{L}_{reg}^{BEV}+\mathcal{L}_{reg}^{z})
    \end{aligned}
\end{equation}
where $\mathcal{L}_{cls}^*$ is the focal loss \cite{lin2017focal} with default hyper-parameters for center classification heads and $\mathcal{L}_{reg}^*$ is the L1 loss for the BBox center offset and rotation regression heads.\\
\noindent{\textbf{Ground Truth Construction.}} Here we take the ground truth construction of BEV heads for instance, and the ground truth of the \textit{z}-axis heads can be similarly formulated. Let $(x,y,z)$ denote the 3D target location, the target 2D center $(c_x,c_y)$ in the BEV is computed as:
\begin{equation}\label{2dcenter}
\begin{aligned}
   c_x&=\frac{x-x_{min}}{v_x},\\
   c_y&=\frac{y-y_{min}}{v_y} \text{.}
\end{aligned}
\end{equation}
The discrete 2D center is defined by $\hat{c_x}=\lfloor c_x\rfloor$ and $\hat{c_y}=\lfloor c_y\rfloor$.
The ground truth of the BEV center classification is $\textbf{y}_{cls}\in \mathbb{R}^{L\times W}$, where the value in location $(i,j)$ becomes 1 if $i=\hat{c_x}$ and $j=\hat{c_y}$, 0 if $(i,j)$ not in the 2D target BBox, otherwise $\frac{1}{\gamma+1}$, where $\gamma$ represents the Euclidean distance between the pixel $(i,j)$ and the discrete target center. 
\begin{table*}[htbp]
\centering
\renewcommand\arraystretch{1.3}
\caption{Comparison among our CorpNet and the state-of-the-art methods on the \textbf{KITTI} datasets. Mean shows the average result weighed by frame numbers. \textbf{Bold} and \underline{underline} denote the best
performance and the second-best performance, respectively.}
\label{kittires}
\vspace{-0.2cm}
\scalebox{0.8}{\begin{tabular}{c|cc|cc|cc|cc|cc}
\hline
                          & \multicolumn{2}{c|}{Car (6424)}                                             & \multicolumn{2}{c|}{Cyclist (308)}                                          & \multicolumn{2}{c|}{Van (1248)}                                             & \multicolumn{2}{c|}{Pedestrian (6088)}                                      & \multicolumn{2}{c}{Mean (14068)}                                            \\ \cline{2-11} 
\multirow{-2}{*}{Methods} & Success                              & Precision                            & Success                              & Precision                            & Success                              & Precision                            & Success                              & Precision                            & Success                              & Precision                            \\ \hline \hline
SC3D \cite{giancola2019leveraging}                      & 41.3                                 & 57.9                                 & 41.5                                 & 70.4                                 & 40.4                                 & 47.0                                 & 18.2                                 & 37.8                                 & 31.2                                 & 48.5                                 \\
SC3D-RPN\cite{zarzar2019efficient}                  & 36.3                                 & 51.0                                 & 43.0                                 & 81.4                                 & -                                    & -                                    & 17.9                                 & 47.8                                 & -                                    & -                                    \\
P2B \cite{qi2020p2b}                      & 56.2                                 & 72.8                                 & 32.1                                 & 44.7                                 & 40.8                                 & 48.4                                 & 28.7                                 & 49.6                                 & 42.4                                 & 60.0                                 \\
MLVSNet \cite{wang2021mlvsnet}                      & 56.0                                 & 74.0                                & 34.3                                 & 44.5                                & 52.0                                & 61.4                                 & 34.1                                 & 61.1                                 & 45.7                                 & 66.6                                 \\
3DSiamRPN \cite{fang20203d}                 & 58.2                                 & 76.2                                 & 36.1                                 & 49.0                                 & 45.6                                 & 52.8                                 & 35.2                                 & 56.2                                 & 46.6                                 & 64.9                                 \\
LTTR \cite{cui20213d}                      & 65.0                                 & 77.1                                 & 66.2                           & 89.9                                 & 35.8                                 & 45.6                                 & 33.2                                 & 56.8                                 & 48.7                                 & 65.8                                 \\
PTT \cite{shan2021ptt}                      & 67.8                                 & 81.8                         & 37.2                                 & 47.3                                 & 43.6                                 & 52.5                                 & 44.9                                 & 72.0                                 & 55.1                                 & 74.2                                 \\
BAT \cite{zheng2021box}                      & 60.5                                & 77.7                                 & 33.7                                 & 45.4                                 & 52.4                                 & 67.0                         & 42.1                                & 70.1                                & 51.2                                & 72.8                                \\
V2B \cite{hui20213d}                      & 70.5                         & 81.3                                 & 40.8                                 & 49.7                                 & 50.1                                 & 58.0                                 & 48.3                                 & 73.5                                 & 58.4                           & 75.2                                 \\
PTTR  \cite{zhou2022pttr}                    & 65.2                                 & 77.4                                 & 65.1                                 & 90.5                           & 52.5                          & 61.8                                 &50.9                           & 81.6                          & 58.4                          &77.8                          \\ 
STNet~\cite{stnet} &\underline{72.1} &\underline{84.0}  &\underline{73.5} &\underline{93.7} &\underline{58.0} &\underline{70.6} &49.9 &77.2 &61.3&80.1\\ 
M2Track~\cite{zheng2022beyond} & 65.5  & 80.8   & 73.2&93.5 &53.8 & \textbf{70.7} &\textbf{61.5}  & \textbf{88.2}   & \underline{62.9}  & \textbf{83.4} \\ 
\hline
\textbf{CorpNet}          & \textbf{73.6}                        & \textbf{84.1}                        & \textbf{74.3}                        & \textbf{94.2}                        & \textbf{58.7}                        &66.5                         & \underline{55.6}                        &\underline{82.4}                        & \textbf{64.5}                        &\underline{82.0}                        \\
\hline
\end{tabular}}
\vspace{-0.23cm}
\end{table*}
\section{Experiment}\label{sec:experiments}
\subsection{Experiment Setups}
\textbf{Implementation details.}
We set both $N_t$ and $N_s$ to 1024 for the input template and search regions by randomly duplicating and discarding points. 
In the encoder, the set abstraction modules from PointNet++ \cite{qi2017pointnet} are simplified into 2 MLP layers to decrease parameters, diminishing the input points to 512, 256, 128 ones, respectively. 
The radius of these layers is set to 0.3, 0.5, and 0.7 meters by default. 
 Parameters of the set abstraction modules and SA modules are shared for the template and search regions. The correlated features after the lateral correlation pyramid are all with 64 channels. In the voxelization process, we set the region $[(x_{min},x_{max}),(y_{min},y_{max}),(z_{min},z_{max})]$ to [(-5.6,5.6),(-3.6,3.6),(-2.4,2.4)] to cover most target points. The voxel size $(v_x,v_y,v_z)$ is set to (0.3,0.3,0.3). For the decoder, three decomposed 3D blocks are stacked before the pooling operations of Eq. \ref{pool}. The two subnetworks include three 2D convolution blocks and three 1D convolution blocks for feature aggregation, respectively. The classification loss has a weight $\lambda_{cls}$ of 1 and the regression loss has a weight $\lambda_{reg}$ of 1. The radius $r$ is set to 2. For all experiments, we use the Adam optimizer with an initial learning rate of 0.001 for training, and the learning rate decays by 0.2 every six epochs by default. It takes about 20 epochs to train our model. In the inference, CorpNet runs at 36 FPS.\\
\textbf{Training and Testing.}
In the training stage, we use the points chosen from the ground truth BBox in the first frame and the ground truth BBox dealt by a random shift from the last frame as the template. The search region is generated with ground truth BBox enlarged by 2 meters plus with the random shift. In the testing stage, we use the points inside the BBox of the first frame and the last predicted BBox as the template. The search region is generated by the last predicted BBox enlarged by 2 meters.
\subsection{Comparison with State-of-the-Art Trackers}
\subsubsection{Comparison on KITTI}
KITTI \cite{kitti} contains 21 training sequences and 29 test sequences. We follow previous works \cite{giancola2019leveraging,zheng2022beyond,zheng2021box} to split the training set into train/val/test splits due to the inaccessibility of the test labels. We use scenes 0-16 for training, scenes 17-18 for validation, and scenes 19-20 for testing.
\begin{table*}[htbp]
\centering
\renewcommand\arraystretch{1.3}
\caption{Comparison among our CorpNet and the state-of-the-art methods on the \textbf{NuScenes} datasets. Mean shows the average result weighed by frame numbers. \textbf{Bold} and \underline{underline} denote the best
performance and the second-best performance, respectively. }
\label{nuscene}
\vspace{-0.3cm}
\scalebox{0.8}{\begin{tabular}{c|cc|cc|cc|cc|cc}
\hline
                          & \multicolumn{2}{c|}{Car (15578)}                                            & \multicolumn{2}{c|}{Bicycle (501)}                                           & \multicolumn{2}{c|}{Truck (3710)}                                           & \multicolumn{2}{c|}{Pedestrian (8019)}                                      & \multicolumn{2}{c}{Mean (27808)}                                            \\ \cline{2-11} 
\multirow{-2}{*}{Methods} & Success                              & Precision                            & Success                              & Precision                             & Success                              & Precision                            & Success                              & Precision                            & Success                              & Precision                            \\ \hline \hline
SC3D \cite{giancola2019leveraging}                     & 24.5                                 & 25.9                                 & 16.6                                 & 18.8                                  & 32.5                           & 30.6                           & 13.8                                 & 14.7                                 & 22.3                                 & 23.2                                 \\
P2B \cite{qi2020p2b}                        & 32.8                                 & 35.2                           & 19.7                                 & 26.6                                  & 16.2                                 & 11.1                                 & 19.2                                 & 26.6                                 & 26.4                                 & 29.3                                 \\
BAT \cite{zheng2021box}                        & 26.5                                 & 28.8                                 & 17.8                                 & 22.8                                  & 16.5                                 & 10.6                                 & 19.4                                 & \underline{28.2}                           & 23.0                                 & 27.9                                 \\
V2B \cite{hui20213d}                        & 32.9                        & 34.5                                 & 20.3                          & 27.5                          & 28.7                                 & 23.8                                 & \underline{20.1}                           & 27.4                                 & 28.4                          &30.9                          \\
STNet~\cite{stnet} &\underline{35.7} &\underline{37.2} &\underline{22.3} &\underline{29.3} &\underline{33.5} &\underline{32.4} &\underline{20.1} &27.8 &\underline{30.7} &\underline{33.7} \\\hline
\textbf{CorpNet}          &\underline{35.0}                        & \textbf{38.4}                        & \textbf{26.9}                        & \textbf{43.5}                         & \textbf{39.7}                        & \textbf{36.3}                        & \textbf{21.3}                        & \textbf{33.6}                        & \textbf{31.8}                        & \textbf{36.8}                        \\
\hline
\end{tabular}}
\vspace{-0.4cm}
\end{table*}

We compare the proposed CorpNet with current state-of-the-art methods, from the pioneering SC3D \cite{giancola2019leveraging} to the most recent STNet \cite{stnet}. As shown in Tab. \ref{kittires}, our CorpNet performs significantly better than other methods on the mean results of four categories. We yield the best results on most categories. We find that our sufficient self aggregations and cross interactions, as well as the correlation pyramid in the encoder, make our CorpNet learn effectively on the data-rare categories like Cyclist and Van. While most previous methods like V2B \cite{hui20213d}, BAT \cite{zheng2021box} and PTT \cite{shan2021ptt} are hard to handle these classes. The second best M2Track \cite{zheng2022beyond} also proposes to change the Siamese-like pipeline by constructing a spatial-temporal point cloud to avoid the Siamese-like backbone and predicts the motion directly. Nevertheless, they still need a motion transformation module to integrate the template information into the search representations, and a two-stage refinement is used to improve the performance. As a result, we achieve better performance than M2Track with a single stage. Besides, compared with V2B \cite{hui20213d} which is the decoder baseline of our method, our proposed encoder and the improved decoder make CorpNet yield better performance in all categories.

\subsubsection{Experiments on NuScenes}
NuScenes~\cite{nuscenes} has 1000 scenes, which are divided into 700/150/150 scenes for train/val/test. Officially, the train set is further evenly split into “train track`` and “train detect`` to remedy overfitting. Following \cite{hui20213d}, we train our model with “train track” split and test it on the val set. 

Note that the NuScenes dataset only labels keyframes and provides official interpolated results for the remaining frames, so there are two configurations for this dataset. The first is from \cite{zheng2021box} which trains and tests both only on the keyframes. The other one is from \cite{hui20213d} which trains and tests on all the frames. These two configurations result in different datasets with different performances. We believe that the motion in key frames is extremely large, which is not in line with the practical applications. Therefore, we follow the second set in this paper.
The methods that follow the second way are compared, and the performance evaluated on the keyframes is reported.

In Tab. \ref{nuscene}, we report the comparison results on NuScenes. The proposed CorpNet exceeds all the competitors under all categories, mostly by large margins. NuScenes is much more challenging than KITTI due to the label scarcity, pervasive distractors, and drastic appearance changes. CorpNet still surpasses other methods on both rigid (e.g., Car) and non-rigid (e.g., Pedestrian) objects, both small (Bicycle) and large (Truck) objects. Besides, our method significantly improves categories (i,e, Bicycle and Truck) with fewer data. The thorough cross-source interactions and the lateral pyramid helps to learn useful features even though the annotation may be incorrect (interpolated labels) on this dataset.

\subsection{Ablation Study}
We comprehensively perform ablation studies to analyze each component in our proposed CorpNet on the Car category of the KITTI dataset. \\
\textbf{Location of SA modules and CA modules.}
We study the influence of the SA and CA modules separately in Tab. \ref{SA_CA}. For both SA and CA, we find that merely adding one module to the last stage has already exceeded all the previous methods. When placing SA and CA on all the three stages, we could obtain the best results. Therefore, we use this configuration in our final CorpNet. These results are intuitive since the representation could be enriched more if the features of all levels are self-aggregated by SA modules. Similarly, the template information could be sufficiently interacted and enhanced if the correlation operations existed in all the stages, yielding better results.\\
\textbf{Number of neighbors in the local attention.}
In Tab. \ref{k_num}, we investigate the influence of the number of neighbors $k$, which determines the considered local neighborhood around each point. We use the same setting for SA and CA modules. In our CorpNet, the best results are obtained when $k$=$32$. When $k$ is smaller, like 8 and 16, the model will have insufficient context for prediction. More specifically, SA may not have enough information to strengthen its features, and CA can not interact with sufficient related points. When $k$ is larger, like 64, the performance will be deteriorated by excessive noise, which is farther and less relevant. The best choice of $k$ is 32 in CorpNet.\\
\noindent\textbf{How important are the correlation pyramid?}
We conduct an ablation study about the correlation pyramid in Tab. \ref{pyramid}. When only using the correlated features of stage 3 without the pyramid structure, the performance drops significantly (2.7\%) compared with the final CorpNet. When fusing the features of stage 2 and stage 3, the results are still worse than fusing all the three stages (-1.0\%). This demonstrates that using the correlation from all stages for the proposed correlation pyramid is important and beneficial to the model's accuracy.\\

\begin{table}[htbp]
\vspace{-0.5cm}
        \centering
          \caption{\centering{Location of SA and CA modules.}}
          \vspace{-0.3cm}
 \scalebox{0.8}{\begin{tabular}{ccc|c|c}
\hline
Stage 1                   & Stage 2                   & Stage 3                  & SA & CA\\ \hline
                          &                           & \checkmark &    71.2/81.9           & 70.8/81.9      \\
                          & \checkmark & \checkmark &  71.3/82.8       & 72.8/83.9           \\
\checkmark & \checkmark & \checkmark &\textbf{73.6}/\textbf{84.1}    &  \textbf{73.6}/\textbf{84.1}   \\ \hline
\label{SA_CA}
\end{tabular}}
\vspace{-0.6cm}
\end{table}
\begin{table}[htbp]
 \vspace{-0.2cm}
        \centering
    \caption{\centering{Number of neighbors $k$.}}
 \vspace{-0.3cm}
\scalebox{0.8}{\begin{tabular}{c|cc}
\hline
k                      & Success              & Precision            \\ \hline
8                      & 69.9                &    81.8            \\
16                     & 71.4                &    81.4     \\
32                     & \textbf{73.6}       &  \textbf{84.1}           \\
64                     &     72.1        &   83.6                   \\ \hline
\end{tabular}}
    \label{k_num}
\vspace{-0.5cm}
\end{table}
\begin{table}[htbp]
\vspace{-0.1cm}
\centering
   \caption{\centering{Influence of the pyramid structure.}}
    \vspace{-0.3cm}
\scalebox{0.8}{\begin{tabular}{ccc|cc}
\hline
Stage 1                   & Stage 2                   & Staget 3               & Success & Precision  \\ \hline
                 &                           & \checkmark &    70.9   &  81.5              \\
                          & \checkmark & \checkmark  &  72.6  &   82.9
                          \\
\checkmark & \checkmark & \checkmark & \textbf{73.6} &\textbf{84.1}     \\ \hline    \\
\end{tabular}}
    \label{pyramid}
    \vspace{-0.6cm}
\end{table}
\noindent \textbf{Is the \textit{z}-axis separation beneficial?}
We now study our contribution in the decoder in Tab. \ref{zaxis}, where 3D Conv is short for 3D We now study our contribution in the decoder in Tab. \ref{zaxis}, where 3D Conv is short for 3D convolution blocks, BEV stands for the 2D prediction head operating on the BEV feature maps, Decomposed 3D Conv represents our proposed decomposed 3D blocks and \textit{z} denotes our standalone prediction head for the \textit{z}-axis features. "3D Conv + BEV`` is actually the original decoder implement of V2B \cite{hui20213d}. The results show that 1) directly using V2B's decoder improves 1.8\% (success) over V2B, which validates the effectiveness of our proposed encoder, 2) only factorizing the 3D convolution block but using only the BEV head is not working, and we attribute to the fact that decomposed 3D convolution first explicitly embeds the \textit{z}-axis information but then pools it will damage the decomposed features and lead to failures.
3) factorizing the 3D convolution block and adding a separate \textit{z}-axis prediction head together is beneficial and further improve the performance with 1.3\% in success, which verifies our initial consideration that the motion pattern of the \textit{x-y} plane and the movement in \textit{z} axis are not exactly the same. 
\begin{table}[htbp]
\vspace{-0.3cm}
\centering
\renewcommand\arraystretch{1.3}
\caption{\textit{z}-axis separation in the decoder.}
\vspace{-0.3cm}
\scalebox{0.8}{
\begin{tabular}{c|cc}
\hline
Configurations                & Success              & Precision            \\ \hline
3D Conv + BEV                 & 72.3             &   81.7             \\
3D Conv + BEV+z    &      70.6       &   82.1   \\
Decomposed 3D Conv + BEV                 &   59.8     &  76.1         \\
Decomposed 3D Conv + BEV+z  &  \textbf{73.6}     &  \textbf{84.1}             \\ \hline
\end{tabular}}
\label{zaxis}
\vspace{-0.5cm}
\end{table}

\subsection{Computational Cost}
We analyze the computational cost in Tab. \ref{runtime}. The results are tested under the same PyTorch platform and a single TITAN RTX GPU on the Car category of KITTI. We can see that the speed of our method is real-time and comparable with V2B, but our CorpNet performs better than V2B with 3.1\% in success. Although BAT and P2B infer faster than our CorpNet, our performance is significantly better than them (+17.4\% and +13.1\% in success). For the training time, our CorpNet, V2B and BAT all converge fast with about 8 hours, while SC3D and P2B require more time. Besides, the parameters and FLOPs of CorpNet are also comparable with other methods. 
\begin{table}[htbp]
\vspace{-0.2cm}
\centering
\renewcommand\arraystretch{1.3}\caption{The Computational cost of different trackers.}
\vspace{-0.2cm}
\label{runtime}
\scalebox{0.8}{\begin{tabular}{c|ccccc}
\hline
Method  & Parameters & FLOPs   & FPS & Training time &Success \\ \hline
SC3D \cite{giancola2019leveraging}     & 6.45 M     & 20.07 G & 6   & $\sim$13 h   &41.3 \\
P2B \cite{qi2020p2b}    & 1.34 M     & 4.28 G  & 48  & $\sim$13 h   &56.2\\
BAT \cite{zheng2021box}      & 1.47 M     & 5.53 G  & 54  & $\sim$8 h   &60.5  \\
V2B \cite{hui20213d}    & 1.35 M     & 5.47 G  & 39  & $\sim$8 h    &70.5 \\
CorpNet & 1.95 M     & 7.02 G    & 36  & $\sim$8 h   &\textbf{73.6}  \\ \hline
\end{tabular}}
\vspace{-0.4cm}
\end{table}
\section{Conclusion}\label{sec:conclusion}
This paper proposes a novel correlation pyramid network (CorpNet) for the 3D single object tracking task, which merges the extractor and matcher of the traditional pipeline to jointly learn the target-aware representation, enabling the extractor and matcher to benefit each other. Particularly, CorpNet integrates multi-level self attentions and cross attentions to enrich the template and search region features, and sufficiently realize their fusion and interaction, respectively. Furthermore, a lateral correlation pyramid structure is designed in the encoder to handle the sparsity problem from the downsampling operations by combining the hierarchical correlated features of all the points that existed in the encoder. The decoder of our method has a new \textit{z}-axis-separated structure, which explicitly learns the movement of the \textit{z} axis and the \textit{x-y} plane. Finally, our method achieves state-of-the-art results on two commonly-used datasets (KITTI and NuScenes). 
{\small
\bibliographystyle{ieee_fullname}
\bibliography{./egbib.bib}
}

\end{document}